\documentclass[letterpaper, 10 pt, conference]{ieeeconf}
\IEEEoverridecommandlockouts
\usepackage{amsmath,amssymb,amsfonts}
\usepackage{graphicx}
\usepackage{array}
\usepackage{units}
\usepackage{makecell}
\usepackage{multirow}
\usepackage[nolist,nohyperlinks]{acronym}
\usepackage{textcomp}
\usepackage{breqn}
\usepackage{mathtools}
\usepackage{verbatim}
\usepackage[export]{adjustbox}
\usepackage{subfig}
\usepackage{mwe}    
\usepackage{tikz}
\usepackage[noadjust]{cite}
\usepackage[colorinlistoftodos]{todonotes}

\newcommand{\etal}{\textit{et al}. }
\newcommand\edit[1]{#1}

\begin{document}

\title{IDOL: A Framework for IMU-DVS Odometry using Lines}
\author{Cedric Le Gentil$^{1,2,\ast}$, Florian Tschopp$^{1,\ast}$, Ignacio Alzugaray$^3$, \\Teresa Vidal-Calleja$^2$, Roland Siegwart$^1$, and Juan Nieto$^1$%
\thanks{$^\ast$ Equal contribution}
\thanks{$^1$Authors are members of the Autonomous Systems Lab, ETH Zurich, Switzerland; {\tt\small \{firstname.lastname\}@mavt.ethz.ch}}%
\thanks{$^2$Authors are members of the Centre for Autonomous Systems, School of Mechanical and Mechatronic Engineering, University of Technology Sydney, Sydney, New South Wales, Australia {\tt\small cedric.legentil@student.uts.edu.au}, {\tt\small teresa.vidalcalleja@uts.edu.au}}%
\thanks{$^3$The author is member of the Vision for Robotics Lab, ETH Zurich, Switzerland; {\tt\small ialzugaray@mavt.ethz.ch}}%
\thanks{This work was partly supported by Siemens Mobility, Germany and the ETH Mobility Initiative under the project \textit{PROMPT}.
}%
\thanks{\textcopyright 2020 IEEE. Personal use of this material is permitted. Permission from IEEE must be obtained for all other uses, in any current or future media, including reprinting/republishing this material for advertising or promotional purposes, creating new collective works, for resale or redistribution to servers or lists, or reuse of any copyrighted component of this work in other works
}
}

\maketitle

\begin{abstract}
In this paper, we introduce IDOL, an optimization-based framework for IMU-DVS Odometry using Lines.
Event cameras, also called Dynamic Vision Sensors (DVSs), generate highly asynchronous streams of events triggered upon illumination changes for each individual pixel. This novel paradigm presents advantages in low illumination conditions and high-speed motions. 
Nonetheless, this unconventional sensing modality brings new challenges to perform scene reconstruction or motion estimation.
The proposed method offers to leverage a continuous-time representation of the inertial readings to associate each event with timely accurate inertial data.
The method's front-end extracts event clusters that belong to line segments in the environment whereas the back-end estimates the system's trajectory alongside the lines' 3D position by minimizing point-to-line distances between individual events and the lines' projection in the image space.
A novel attraction/repulsion mechanism is presented to accurately estimate the lines' extremities, avoiding their explicit detection in the event data.
The proposed method is benchmarked against a state-of-the-art frame-based visual-inertial odometry framework using public datasets. The results show that IDOL performs at the same order of magnitude on most datasets and even shows better orientation estimates. These findings can have a great impact on new algorithms for DVS.
\end{abstract}

\section{Introduction}
In mobile robotics, having an understanding of the environment and the system's position in it is essential \cite{RolandSiegwart2011}. While being cost-effective, visual \ac{slam} and \ac{vo} have been shown to achieve high accuracy and robustness in many applications \cite{Mur-Artal2015ORB-SLAM:System,Cadena2016,Engel2018DirectOdometry}. By adding an \ac{imu}, the accuracy and robustness of \ac{vo} can be further improved \cite{Bloesch2015, Leutenegger2015, Qin2018VINS-Mono:Estimator, Tschopp2019ExperimentalVehiclesb}. However, there are still scenarios which are challenging for visual-inertial systems such as under very fast motions or in scenes with \ac{hdr} of illumination. \\
\acp{dvs}, also called event-based cameras, are new sensor types that have a huge potential in addressing aforementioned limitations due to their extremely high temporal resolution and their \ac{hdr} operative range \cite{Lichtsteiner2008, Brandli2014}. In contrast to traditional frame-based cameras, which periodically output intensity values for all pixels, these sensors output event tuples $e = \left\{t,x,y,p\right\}$, where $t$ is the timestamp of the event, $x,y$ are the image coordinates, and $p$ is the event polarity. Such an event is triggered only in case a pixel's intensity change is larger than a threshold, with $p$ being the direction of that change.
Consequently, in a static scene, events are only generated when the camera is in motion. 
The events form an asynchronous stream of data that provides reliable information even in the presence of fast motion.
As of today, most developments in the event-based literature focused on techniques to aggregate the event data into key-frames in order to apply or adapt conventional frame-based \ac{vo} algorithms.
To fully leverage the potential of this novel type of modality, new algorithms and ego-motion estimation paradigms need to be developed.

\begin{figure}
    \newcolumntype{C}[1]{>{\centering\let\newline\\\arraybackslash\hspace{0pt}}m{#1}}
    \def\colsize{3.3cm}
    \centering
    \begin{tabular}{C{\colsize} C{\colsize}}
        \includegraphics[clip, width=\colsize,trim=0cm 0cm 0cm 0cm]{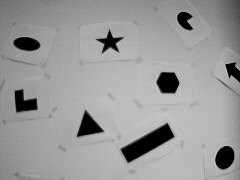}
        &
        \includegraphics[clip, width=\colsize,trim=0cm 0cm 0cm 0cm]{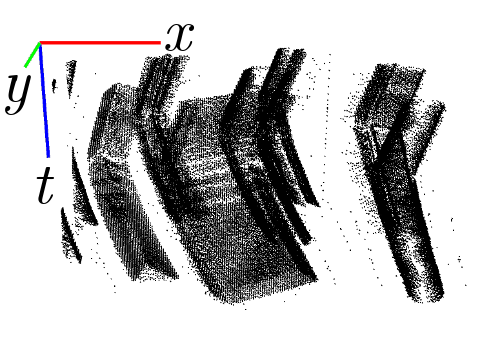}
        \\
        \small(a) Greyscale image of environment \cite{Mueggler2017}
        &
        \small(b) Spatio-temporal view of raw stream of events
        \\
        \includegraphics[clip, width=\colsize,trim=0cm 0cm 0cm 0cm]{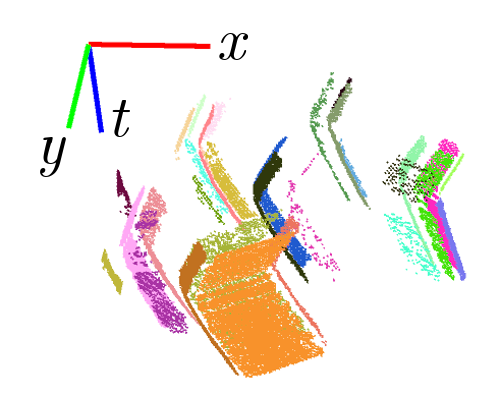}
        &
        \includegraphics[clip, width=\colsize,trim=0cm 0cm 0cm 0cm]{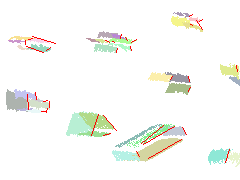}
        \\
        \small(c) Event clustering of line segments
        &
        \small(d) Estimated lines, in red, projected in the image space
    \end{tabular}
    \caption{The proposed method, IDOL, estimates the system's ego-motion based on the segmentation and position estimation of line segments in the environment.}
    \label{figure:teaser}
\end{figure} %
Most of the state-of-the-art work in both traditional and event-based \ac{vo} rely on the tracking of point-like features over time.
Those feature-tracks are used in a filter or optimization-based back-end to estimate the camera motion and the 3D location of the observed points. Human-made structures, however, are built with regular geometric shapes such as lines, making point landmarks not necessarily the best representation for visual tracking in all the scenarios.
\edit{In its current state, our front-end does not offer enough robustness to address the \texttt{boxes} scenarios of \cite{Mueggler2017} because of their very high level of texture in the scene.}

In this paper, we investigate the potential of directly using asynchronous events without any frame-like accumulation, for ego-motion estimation.
A key element is the use of a continuous-time representation of the inertial data to constrain the system pose at any time without relying on any motion model.
By tightly coupling \ac{imu} and event data together via the generation of inertial information at each event's timestamp, a \ac{vio} formulation that addresses and leverages the data asynchronism rigorously is achieved.
Furthermore, instead of using traditional point features, we represent the environment using line segments.
We introduce a new pipeline, \ac{idol}, that detects line features in the event data and performs \ac{vio} by individually considering asynchronous events in a batch-optimization framework.

The remainder of the paper is organized as follows: In Section~\ref{sec:rel_work}, a summary of related work in event-based motion estimation, continuous state and measurement representation is presented. Section~\ref{sec:method} provides an overview of the proposed framework \ac{idol} while Section~\ref{sec:backend} and Section~\ref{sec:frontend} gives more details about the front-end and back-end, respectively. In Section~\ref{sec:experiments}, \ac{idol} is evaluated on public datasets and compared to state-of-the-art in traditional frame-based \ac{vio}. Finally, Section~\ref{sec:conclusions} provides a conclusion with an outlook on future work.

\section{Related Work} \label{sec:rel_work}

\subsection{Event-based motion estimation}
The major differences between the traditional frame-based and event-based vision make the latter an especially appealing sensing modality for the task of \ac{vo} in challenging scenarios where the performance of traditional imagers is compromised such as in \ac{hdr} scenes or under fast camera motions. Despite the relatively recent interest of the community in event cameras, we can already identify several works on \ac{vo} employing event cameras over the last years, starting from the first 2D \ac{slam} approach by Weikersdorfer \etal \cite{weikersdorfer:ICVS2013:event-slam}. Years later, Kim \etal \cite{kim:ECCV2016:slam} achieved the estimation of the 6-\ac{dof} of the camera pose on generic 3D scenes using probabilistic filtering. Rebecq \etal \cite{rebecq:RAL2017:evo} proposed a parallel mapping and tracking approach that iteratively co-localize the pose of the camera against a local map of edges represented by a voxel grid.

Aforementioned approaches avoid the explicit definition of features at expenses of being relatively demanding in terms of computational resources. Modern event-based \ac{vio} pipelines such as \cite{zhu:CVPR2017:evio,rebecq:BMVC2017:viokeyframe,vidal:ICRA2018:ultimate-slam}, however, rely on detection and tracking of corners employing intermediate image-like representations from the accumulation of events. These approaches make use of \ac{imu}s, profiting from their high-rate of inertial measurements and making them an appealing type of sensor to be combined with  event cameras.

While the bulk of event-based \ac{vo} approaches still rely on the traditional concept of key-frames, it is only natural that continuous-time approaches would emerge. One of the seminal works is introduced by Mueggler \etal \cite{mueggler:RSS2015:continuous}, in which the state of the camera is estimated associating events with line segments evaluated both in simulation and real-world experiments using fiducial markers detected using intensity images. The same authors later expand their approach to consider inertial measurements in \cite{mueggler:TRO2018:continuous-vi}.

Over the years, significant efforts in the community have been dedicated to the definition of reliable visual features for event data.
It has led to the development of mainly corner detection and tracking approaches ({\it e.g.} \cite{vasco:IROS2016:harris, manderscheid:CVPR2019:speed-invariant, alzugaray:RAL2018:corner-detector,alzugaray:3DV2018:ace}), while the progress of line-based features has been comparatively less notable. Among other works, we could highlight the approaches proposed by Br\"andli \etal \cite{brandli:EBCCSP2016:elised}, detecting and tracking line clusters in an event-by-event fashion, and Everding \etal \cite{everding:fnbot2018:line-tracking}, whose propose a  plane fitting approach on the event stream based on principal component analysis to track and cluster lines.

The proposed approach takes inspiration from previous continuous-time formulations and their integration with \ac{imu} measurements as in  \cite{mueggler:RSS2015:continuous} and  \cite{mueggler:TRO2018:continuous-vi}. However, we also draw concepts from event-driven line tracking and clustering approaches such as the methods described in \cite{brandli:EBCCSP2016:elised} and \cite{everding:fnbot2018:line-tracking}, avoiding the need for supplementary sensing modalities as frame-based images and operating directly on the asynchronous event stream.

\subsection{Continuous state and measurement representation}

The asynchronism of event cameras represents a major technical challenge for state estimation.
While originally focused on discrete-time estimation, the use of rolling-shutter-like sensors (lidar, rolling-shutter camera, etc.), as well as multi-sensor platforms, lead the robotics community to develop continuous-time state estimation theory and methods.
A large number of frameworks assume motion-models, often constant velocity, to interpolate the state variables in between discrete estimation timestamps \cite{Hong2010,Bosse2009}.
In \cite{Furgale2012}, Furgale \etal introduce a fully continuous framework that considers the state as being the linear combination of temporal basis functions.
Anderson and Barfoot \cite{Anderson2015} present a probabilistic approach to efficiently infer the state variables using \ac{gp} regression over a discrete maximum a posteriori estimation.\\
A different paradigm is presented in \cite{LeGentil2018, LeGentil2019} where \acp{gp} are used as continuous representations of the inertial data allowing the characterization of the system's pose in a continuous manner while relying on a discrete state.
This is the approach employed in the proposed method with the use of \acp{gpm} originally presented in \cite{LeGentil2020}.

\section{Method overview} \label{sec:method}

The proposed method aims at estimating the ego-motion of an event-based visual-inertial system.
To this end, line segments are detected in the event data provided by the camera, and both the system's trajectory and the position of the 3D lines are simultaneously estimated.
Addressing the asynchronicity of the event stream, a continuous representation of the inertial data is used to associate each event with inertial measurements.
The state is then estimated by means of a discrete-state batch on-manifold optimization that accounts for the events individually.

\subsection{Problem formulation}
\def\a{a}
\def\b{b}
\def\e{e}
\def\m{m}
\def\mm{m-1}
\def\mp{m+1}
\def\i{i}
\def\j{j}
\def\l{l}
\def\q{q}
\newcommand\eventtime[1]{t_{#1}}
\newcommand\frametime[1]{\tau_{#1}}
\newcommand\imutime[1]{\mathfrak{t}_{#1}}
\newcommand\refframe[2]{\mathfrak{F}_{#1}^{#2}}
\def\rc{\mathbf{R}_I^C}
\def\pc{\mathbf{p}_I^C}
\def\costfunction{{C}}
\def\state{\mathcal{S}}
\newcommand\event[1]{\mathbf{e}^{#1}}
\newcommand\eventcomponent[2]{{e}^{#1}_{#2}}
\def\nbframe{M}
\def\nbimu{Q}
\def\nbline{L}
\def\windowsize{N}
\newcommand\rot[2]{\mathbf{R}_{#1}^{#2}}
\newcommand\pos[2]{\mathbf{p}_{#1}^{#2}}
\newcommand\vel[2]{\mathbf{v}_{#1}^{#2}}
\newcommand\biasacccorrection[1]{\hat{\mathbf{b}}_{f}^{#1}}
\newcommand\biasgyrcorrection[1]{\hat{\mathbf{b}}_{\omega}^{#1}}
\newcommand\biasaccprior[1]{\bar{\mathbf{b}}_{f}^{#1}}
\newcommand\biasgyrprior[1]{\bar{\mathbf{b}}_{\omega}^{#1}}
\newcommand\biasacc[1]{\mathbf{b}_{f}^{#1}}
\newcommand\biasgyr[1]{\mathbf{b}_{\omega}^{#1}}
\def\imu{I}
\def\cam{C}
\def\world{W}
\newcommand\projectfunction[1]{\pi(#1)}
\newcommand\properacc[1]{\tilde{\mathbf{f}}_{#1}}
\newcommand\rawgyr[1]{\tilde{\boldsymbol{\omega}}_{#1}}
\newcommand\lineend[3]{{\mathbf{l}_{#3}}^{#2}_{#1}}
\newcommand\linepixel[2]{\mathbf{d}^{#1}_{#2}}
\newcommand\linestate[2]{{\mathbf{L}}^{#2}_{#1}}
\newcommand\imuresidual[1]{\mathbf{r}_{\imu}^{#1}}
\newcommand\imuresidualp[1]{\mathbf{r}_{\imu_p}^{#1}}
\newcommand\imuresidualv[1]{\mathbf{r}_{\imu_v}^{#1}}
\newcommand\imuresidualr[1]{\mathbf{r}_{\imu_r}^{#1}}
\newcommand\biasaccresidual[1]{\mathbf{r}_{f}^{#1}}
\newcommand\biasgyrresidual[1]{\mathbf{r}_{\omega}^{#1}}
\newcommand\segresidual[1]{r_{l}^{#1}}
\newcommand\attractionresidual[1]{r_{a}^{#1}}
\newcommand\splittingresidual[1]{r_{s}^{#1}}
\newcommand\sigmaimu[1]{\Sigma{\mathbf{r}_{\imu}^{#1}}}
\newcommand\sigmabiasacc[1]{\Sigma_{\mathbf{r}_{f}^{#1}}}
\newcommand\sigmabiasgyr[1]{\Sigma_{\mathbf{r}_{\omega}^{#1}}}
\newcommand\sigmaseg[1]{\Sigma_{\mathbf{r}_{l}^{#1}}}
\newcommand\sigmaattraction[1]{\Sigma_{\mathbf{r}_{a}^{#1}}}
\newcommand\sigmasplitting[1]{\Sigma_{\mathbf{r}_{s}^{#1}}}
\def\assoset{\mathcal{A}}
\def\asso{\alpha^{\i}}
\newcommand\transform[2]{\mathbf{T}_{#1}^{#2}}

Let us consider a rigidly mounted event-camera and a 6-DoF \ac{imu}.
The camera and \ac{imu} reference frames at time $\eventtime{\i}$ are respectively noted $\refframe{\cam}{\eventtime{\i}}$ and $\refframe{\imu}{\eventtime{\i}}$.
The relative transformation from $\refframe{\imu}{\eventtime{\i}}$ to $\refframe{\cam}{\eventtime{\i}}$ is characterized by the rotation matrix $\rc$  and the translation vector $\pc$.
Homogeneous transformation will be used for the rest of the paper, therefore rotation matrices and translations/positions will be associated with $4\times4$ transformation matrices with the same combination of subscripts and superscripts,
\begin{equation}
    \transform{a}{b} = \begin{bmatrix}\rot{a}{b}&\pos{a}{b}\\\mathbf{0}^\top & 1\end{bmatrix} \text{ and }
        {\transform{a}{b}}^{-1} = \begin{bmatrix}{\rot{a}{b}}^\top&-{\rot{a}{b}}^\top\pos{a}{b}\\\mathbf{0}^\top & 1\end{bmatrix}.
\end{equation}

The 6-\ac{dof} \ac{imu} acquires proper acceleration $\properacc{\imutime{\q}}$ and angular velocity $\rawgyr{\imutime{\q}}$ measurements at time $\imutime{\q}$ ($\q = 1,\cdots,\nbimu$).
These readings are combined together into \acp{gpm} \cite{LeGentil2020} assuming known \ac{imu} biases.
Section~\ref{section:gpm} presents a brief introduction on \acp{gpm} and the mechanism of post-integration bias correction.
The event-camera data is collected as an asynchronous stream of events $\event{\i} = \begin{bmatrix} \eventcomponent{\i}{x}&\eventcomponent{\i}{y} \end{bmatrix}$ at time $\eventtime{\i}$, with $\eventcomponent{\i}{x}$ and $\eventcomponent{\i}{y}$ being the pixel coordinates in the image space.
Note that in this work, we do not consider the polarity of the events.
This stream is arbitrarily organised in $\nbframe$ windows of $\windowsize$ events.
A 3D point is projected into the image space with the function $\projectfunction{\bullet}$ according to a pinhole camera-model.

The system's \ac{imu} orientation $\rot{\world}{\frametime{\m}}$, position $\pos{\world}{\frametime{\m}}$, and velocity $\vel{\world}{\frametime{\m}}$ are estimated at the timestamp $\frametime{\m}$ of the first event of each window ($\m = 1,\cdots,\nbframe$) with respect to the fixed world frame $\refframe{\world}{}$.
The proposed method also estimates corrections, $\biasacccorrection{\m}$ and $\biasgyrcorrection{\m}$, to the bias priors used during the preintegration.

Along the sensor's trajectory, events are clustered into segments that correspond to 3D lines in the environment (front-end).
The accumulation of these event-line associations form the set $\assoset$.
The positions of these lines are estimated simultaneously to the \ac{imu} pose and velocities mentioned above.
Each line is parameterized by two 3D points as 
\begin{equation}
    \linestate{\world}{\l} = \begin{bmatrix} \lineend{\world}{\l}{\a}{}^{\top} & \lineend{\world}{\l}{\b}{}^{\top} \end{bmatrix},
    \label{eq:line}
\end{equation}
with $\l = 1,\cdots,\nbline$.
While this representation over-parameterizes 3D lines, it allows the proposed method to fit the lines' extremities to the actual line segments through a mechanism of attraction/repulsion detailed in Section~\ref{section:linefactors}.

The proposed method estimates the state $\state = (\rot{\world}{\frametime{1}},\cdots,\rot{\world}{\frametime{\nbframe}}$, $\pos{\world}{\frametime{2}},\cdots,\pos{\world}{\frametime{\nbframe}}$, $\vel{\world}{\frametime{2}},\cdots,\vel{\world}{\frametime{\nbframe}}$, $\biasacccorrection{\frametime{1}},\cdots,\biasacccorrection{\frametime{\nbframe}}$, $\biasgyrcorrection{\frametime{1}},\cdots,\biasgyrcorrection{\frametime{\nbframe}}$, $\linestate{\world}{1},\cdots,\linestate{\world}{\nbline})$ using maximum likelihood estimation that corresponds to the minimization of the cost function $\costfunction$:
\begin{align}
    &\state^* = \underset{\state}{\text{argmin}}\; \costfunction(\state),
    \nonumber
    \\
    \costfunction(\state) = &\sum_{\asso \in \assoset}^{} \Big( \lVert \segresidual{\asso} \lVert^2_{\sigmaseg{\asso}} + \lVert \splittingresidual{\asso} \lVert^2_{\sigmasplitting{\asso}} \Big) +
    \sum_{\l = 1}^{\nbline} \lVert \attractionresidual{\l} \lVert^2_{\sigmaattraction{\l}} +
    \nonumber
    \\
    &\sum_{\m = 1}^{\nbframe-1} \Big( \lVert \biasaccresidual{\m} \lVert^2_{\sigmabiasacc{\m}} + \lVert \biasgyrresidual{\m} \lVert^2_{\sigmabiasgyr{\m}} + \lVert \imuresidual{\m} \lVert^2_{\sigmaimu{\m}}\Big),
    \label{eq:costfunction}
\end{align}
with $\segresidual{}$ being event-to-line distances for each event-line association in $\assoset$, $\splittingresidual{}$ and $\attractionresidual{}$ being repulsion and attraction forces between each of the lines' extremities, respectively, $\biasaccresidual{}$ and $\biasgyrresidual{}$ constraints on the IMU biases random-walk, and $\imuresidual{}$ being direct pose and velocity constraints between two consecutive timestamps of the estimated trajectory based on the IMU readings.
These factors (back-end) are detailed in Section~\ref{sec:backend}.
Note that $\Sigma_\bullet$ is the covariance matrix of the variable $\bullet$.

\subsection{Gaussian Preintegrated Measurement}

\label{section:gpm}

\newcommand\dpos[2]{\Delta\mathbf{p}_{#1}^{#2}}
\newcommand\dvel[2]{\Delta\mathbf{v}_{#1}^{#2}}
\newcommand\drot[2]{\Delta\mathbf{R}_{#1}^{#2}}
\def\gravity{\mathbf{g}}

The \acp{gpm} \cite{LeGentil2020} rely on the use of \ac{gp} regression and linear operators to preintegrate (analytically for the position and velocity parts) the inertial measurements between any given timestamps.
Therefore, given the state $\state$, the pose and velocity of the system can be queried at $\eventtime{\i}$ as
\begin{align}
    \pos{\world}{\eventtime{\i}} &= \pos{\world}{\frametime{\m}} + (\eventtime{\i} - \frametime{\m})\vel{\world}{\frametime{\m}} +
    (\eventtime{\i} - \frametime{\m})^2\frac{\gravity}{2} + \rot{\world}{\frametime{\m}}\dpos{\frametime{\m}}{\eventtime{\i}}
    \nonumber
    \\
    \vel{\world}{\eventtime{\i}} &= \vel{\world}{\frametime{\m}} + (\eventtime{\i} - \frametime{\m})\gravity + \rot{\world}{\frametime{\m}}\dvel{\frametime{\m}}{\eventtime{\i}}
    \nonumber
    \\
    \rot{\world}{\eventtime{\i}} &= \rot{\world}{\frametime{\m}}\drot{\frametime{\m}}{\eventtime{\i}}
    \label{eq:continuous_pose}
\end{align}
where $\gravity$ is the known gravity vector in $\refframe{\world}{}$, and $\dpos{\frametime{\m}}{\eventtime{\i}}$, $\dvel{\frametime{\m}}{\eventtime{\i}}$, and $\drot{\frametime{\m}}{\eventtime{\i}}$ the position, velocity and rotation \acp{gpm}, respectively.
\edit{To prevent overly verbose notation, the superscript ${}^{\eventtime{\i}}$ and ${}^{\frametime{\m}}$ refer to the \ac{imu} frame at time $\eventtime{\i}$ and $\frametime{\m}$.}

These pseudo-measurements are functions of the acceleration biases $\biasacc{}$, and gyroscope biases $\biasgyr{}$.
Unfortunately, these values are not accurately known at the time of preintegration.
Consequently, as in \cite{Lupton2012}, the \ac{gpm} approach uses the first-order Taylor expansion of each of the preintegrated measurements $\dpos{\frametime{\m}}{\eventtime{\i}}$, $\dvel{\frametime{\m}}{\eventtime{\i}}$, and $\drot{\frametime{\m}}{\eventtime{\i}}$, and assumes the biases are individually constant in each of the $\nbframe$ windows.
The expansion is based on the approximation that $\biasacc{\m} \approx \biasaccprior{\m} + \biasacccorrection{\m}$ and $\biasgyr{\m} \approx \biasgyrprior{\m} + \biasgyrcorrection{\m}$, where $\biasaccprior{\m}$ and $\biasgyrprior{\m}$ are the prior knowledge of the biases used to compute the \acp{gpm}, and $\biasacccorrection{\m}$ and $\biasgyrcorrection{\m}$ are their first-order correction.

\section{Back-end} 
\label{sec:backend}

This section describes the different elements of the cost function $\costfunction(\state)$ presented in Equation~\eqref{eq:costfunction}.

\subsection{Event-to-line factors}

The event-to-line factors correspond to the point-to-line distances between events in the image space and the image projection of the associated 3D lines.
Let us consider an event-line association $\asso = \{ \event{\i}, \eventtime{\i}, \linestate{\world}{\l}\}$.
The \edit{projections $\linepixel{\eventtime{\i}}{\a_{\l}}$ and $\linepixel{\eventtime{\i}}{\b_{\l}}$} of the line extremities \edit{$\lineend{\world}{\l}{\a}$ and $\lineend{\world}{\l}{\b}$ into} the camera image \edit{at time $\eventtime{\i}$ are obtained} using Equation~\eqref{eq:continuous_pose} and the extrinsic calibration $\transform{\imu}{\cam}$
\begin{align}
    \linepixel{\eventtime{\i}}{\a_{\l}} = \projectfunction{{\transform{\imu}{\cam}}^\top {\transform{\world}{\eventtime{\i}}}^\top \lineend{\world}{\l}{\a}}\text{ and }
    \linepixel{\eventtime{\i}}{\b_{\l}} = \projectfunction{{\transform{\imu}{\cam}}^\top {\transform{\world}{\eventtime{\i}}}^\top \lineend{\world}{\l}{\b}}.
\end{align}

The point-to-line distance residual $\segresidual{\asso}$ is then equal to
\begin{align}
    \segresidual{\asso} = \textstyle\frac{\lVert (\event{\i} - \linepixel{\eventtime{\i}}{\a_{\l}}) \times
        (\linepixel{\eventtime{\i}}{\b_{\l}} - \linepixel{\eventtime{\i}}{\a_{\l}}) \lVert}
        {\lVert \linepixel{\eventtime{\i}}{\b_{\l}} - \linepixel{\eventtime{\i}}{\a_{\l}}\lVert}
\end{align}

\subsection{Line attraction and repulsion factors}
\label{section:linefactors}

\newcommand\linedist[1]{d^{#1}}

\begin{figure}
    \centering
    \begin{tikzpicture}
        \node[anchor=south west] at (0,0) {\includegraphics[clip, trim=0cm 0cm 0cm 0cm, width=7.5cm]{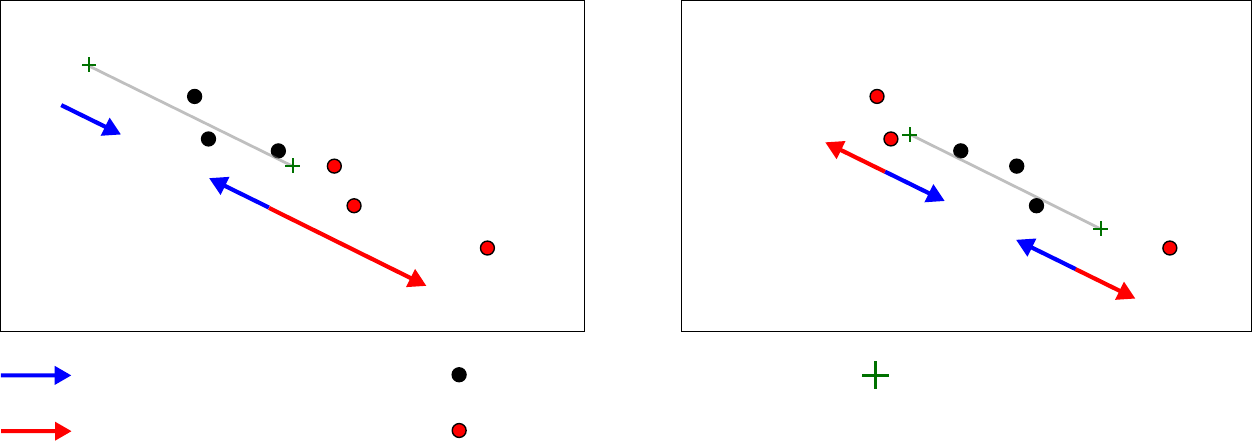}};
        \node[anchor=south] at (6.0,2.7) {\footnotesize After optimization };
        \node[anchor=south] at (2.0,2.7) {\footnotesize Before optimization };
        \node[anchor=west] at (0.6,0.58) {\footnotesize Attraction force};
        \node[anchor=west] at (0.6,0.18) {\footnotesize Repulsion force};
        \node[anchor=west] at (3.0,0.58) {\footnotesize Event};
        \node[anchor=west] at (3.0,0.18) {\footnotesize Event involved in repulsion force};
        \node[anchor=west] at (5.45,0.54) {\footnotesize Line extremity};
    \end{tikzpicture}
    \caption{Illustration of the attraction/repulsion mechanism used to fit the line extremities to the observed segment, as well as to constrain the over-parameterized two-point line representation. The estimated 3D lines extremities are projected into the image space. The extremities are subject to a constant attraction force toward one another. The events that are ``outside" the  line projection induce a repulsion force that push the extremities apart. After optimization, the estimated line fits the actual segments.}
    \label{fig:attraction_repulsion}
\end{figure}
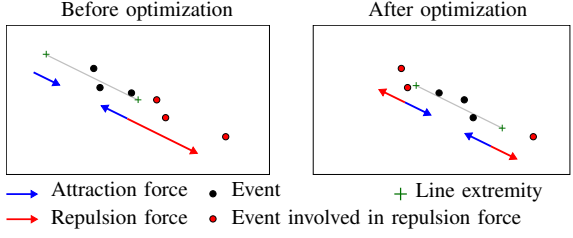

The two-3D-point representation in Equation~\eqref{eq:line} is an over-parameterization of an infinite 3D line, and the event-to-line factors do not fully define the position of the line points.
In other words, without additional constraints, there is ambiguity on the estimated state as an infinite number of point pairs can characterize the same line.
\edit{Having extra unconstrained \acp{dof} may lead to failure of the optimization process.}
To address this issue, we introduce an attraction/repulsion strategy that fits the line extremities to the cluster of events, constraining, therefore, the extra \acp{dof} of the two-point line parameterization.
Intuitively, the two components of this mechanism can be thought as of, on the one side, an inherent attraction force between the line extremities, and on the other, a set of forces generated by the events pushing the extremities apart as illustrated in Figure~\ref{fig:attraction_repulsion}.

Formally, the attraction component is implemented as a residual equal to the square-root of the pixel-distance between the line extremities after projection into the image space at $\frametime{\m}$, the time of the window during which the line has been first observed:
\begin{align}
    \attractionresidual{\l} = \sqrt{\lVert\linepixel{\frametime{\m}}{\b_{\l}} - \linepixel{\frametime{\m}}{\a_{\l}} \lVert}.
\end{align}

The repulsion component is generated by the points around the extremities of the line.
Given an event-line association $\asso$ the position $\linedist{\i}$ of $\event{\i}$ along the line $\linepixel{\eventtime{\i}}{\a_{\l}}$/$\linepixel{\eventtime{\i}}{\b_{\l}}$ is computed as
\begin{align}
    \linedist{\i} = \textstyle\frac{(\event{\i} - \linepixel{\eventtime{\i}}{\a_{\l}})^\top
        (\linepixel{\eventtime{\i}}{\b_{\l}} - \linepixel{\eventtime{\i}}{\a_{\l}})}
        {\lVert \linepixel{\eventtime{\i}}{\b_{\l}} - \linepixel{\eventtime{\i}}{\a_{\l}}\lVert}.
\end{align}
The events that are ``outside" the line lead to residuals equal to the distance along the line to the closest extremity: 
\begin{align}
    \splittingresidual{\asso} = \left\{
        \begin{array}{ll}
            \linedist{\i} & \mbox{if } \linedist{\i} < 0
            \\
            \lVert \linepixel{\eventtime{\i}}{\b_{\l}} - \linepixel{\eventtime{\i}}{\a_{\l}} \lVert - \linedist{\i} & \mbox{if } \linedist{\i} > \lVert \linepixel{\eventtime{\i}}{\b_{\l}} - \linepixel{\eventtime{\i}}{\a_{\l}} \lVert
            \\0 & \mbox{otherwise.}
        \end{array} \right.
\end{align}

Note that this approach does not require the front-end to extract line extremities among the event data.
\edit{The events that are projected outside the estimated line segments automatically generate repulsion forces.
Therefore, the positions of the line extremities are best estimated when solely a small number of events generate repulsion forces.}
\edit{
As repulsion forces are invariant to the lines' length but correlated to the number of events involved, the attraction forces also need to be somewhat length-invariant to prevent the need for any additional balancing mechanism between the forces.
Intuitively, in least-square optimization problems, the Jacobians of the cost function represent forces that constrain the state estimate.
Consequently, the attraction forces are made length-invariant by the use of the square root in  $\attractionresidual{\m}$, making the magnitude of the attraction force ``constant" (preventing long line estimates to be ``squashed" if Euclidean norms were directly used).

}



\subsection{IMU and bias factors}

\newcommand\dtframe[1]{\Delta\frametime{#1}}

The inertial measurements are used to constrain the system's trajectory from one window to the next.
The IMU factors' residual $\imuresidual{\m} = \begin{bmatrix}\imuresidualp{\m} &\imuresidualv{\m} &\imuresidualr{\m} \end{bmatrix}$ are obtained by manipulating Equation~\eqref{eq:continuous_pose}, 
\begin{align}
    \imuresidualp{\m} &= {\rot{\world}{\frametime{\m}}}^\top(\pos{\world}{\frametime{\mp}} - \pos{\world}{\frametime{\m}} - \dtframe{\m}\vel{\world}{\frametime{\m}} -
    \dtframe{\m}^2\frac{\gravity}{2}) - \dpos{\frametime{\m}}{\frametime{\mp}}
    \nonumber
    \\
    \imuresidualv{\m} &= {\rot{\world}{\frametime{\m}}}^\top(\vel{\world}{\frametime{\mp}} - \vel{\world}{\frametime{\m}} - \dtframe{\m}\gravity) - \dvel{\frametime{\m}}{\frametime{\mp}}
    \nonumber
    \\
    \imuresidualr{\m} &= \text{Log}({\drot{\frametime{\m}}{\frametime{\mp}}}^\top {\rot{\world}{\frametime{\m}}}^\top \rot{\world}{\frametime{\mp}})
\end{align}
with $\dtframe{\m}=\frametime{\mp}-\frametime{\m}$, and $\text{Log}(\bullet)$ the mapping from $SO(3)$ (rotation matrix) to $\mathfrak{so}(3)$ (axis-angle).

The proposed method models the temporal evolution of the biases with a Brownian motion.
Consequently, the bias factors' residuals are defined as
\begin{align}
    \biasaccresidual{\m} &= \biasaccprior{\mp} + \biasacccorrection{\mp} - \biasaccprior{\m} - \biasacccorrection{\m}
    \nonumber
    \\
    \biasgyrresidual{\m} &= \biasgyrprior{\mp} + \biasgyrcorrection{\mp} - \biasgyrprior{\m} - \biasgyrcorrection{\m}.
\end{align}

\section{Front-end} \label{sec:frontend}

\newcommand\point[1]{\mathbf{x}^{#1}}
\newcommand\normal[1]{\mathbf{n}^{#1}}
\def\normalthr{n_{\text{thr}}}
\def\ptolthr{e_{\text{thr}}}

Similarly to \cite{everding:fnbot2018:line-tracking}, the proposed method considers the stream of events as 3D information, where the first two components are the events' coordinates in the image space, and the third coordinate is the events' timestamps arbitrarily normalized: $\point{\i} = \begin{bmatrix} \eventcomponent{\i}{x}& \eventcomponent{\i}{y} & \eventtime{\i}/c \end{bmatrix}{}^\top$.
\edit{The value of $c$ is chosen according to the average level of texture in the scene.}
The front-end consists of clustering events that are triggered by the same physical line in the environment according to the premise that 3D lines translate to locally planar patches in the 3D spatio-temporal representation of the event stream.
At the moment, the presented approach does not aim for real-time operation, and thus it uses windows of $\windowsize$ events to perform the event clustering.
The event data in each window can be seen as a point cloud with the 3D-points' coordinates defined as $\point{\i}$.
Normal vectors are estimated for each of the points based on the eigenvectors of the covariance matrix built with the neighbouring points.
Points are considered neighbours if their Euclidean distance in the 3D spatio-temporal space is under a certain given threshold.
\edit{Note that the proposed method does not explicitly fit planes to the 3D event data throughout the windows, but computes the normals based on a local neighbourhood of events.}
The $x$ and $y$ components of the normals are normalized to unit vectors $\normal{\i}$.

\begin{figure}
    \newcolumntype{C}[1]{>{\centering\let\newline\\\arraybackslash\hspace{0pt}}m{#1}}
    \def\colsize{1.8cm}
    \centering
    \setlength{\tabcolsep}{1pt}
    \begin{tabular}{C{\colsize}| C{\colsize}| C{\colsize}| C{\colsize}}
        \includegraphics[clip, width=\colsize,trim=0cm 0cm 0cm 0cm]{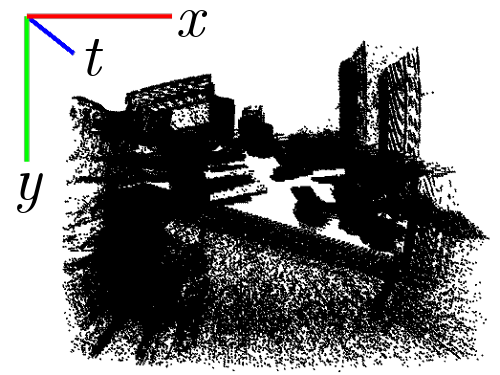}
        &
        \includegraphics[clip, width=\colsize,trim=0cm 0cm 0cm 0cm]{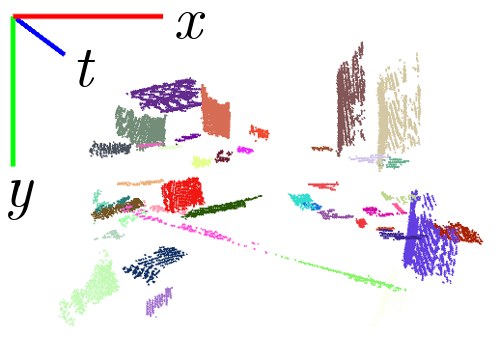}
        &
        \includegraphics[clip, width=\colsize,trim=0cm 0cm 0cm 0cm]{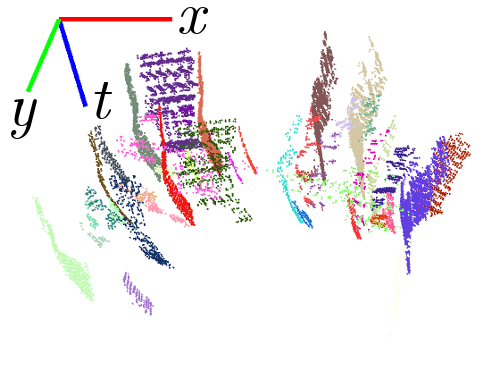}
        &
        \includegraphics[clip, width=\colsize,trim=0cm 0cm 0cm 0cm]{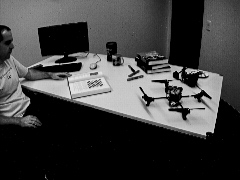}
        \\
        \hline
        \includegraphics[clip, width=\colsize,trim=0cm 0cm 0cm 0cm]{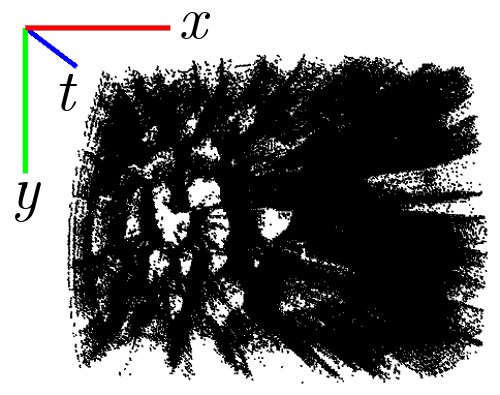}
        &
        \includegraphics[clip, width=\colsize,trim=0cm 0cm 0cm 0cm]{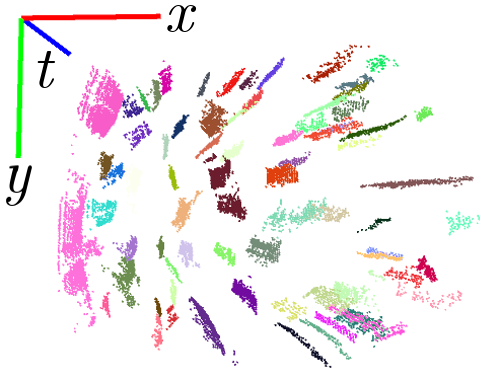}
        &
        \includegraphics[clip, width=\colsize,trim=0cm 0cm 0cm 0cm]{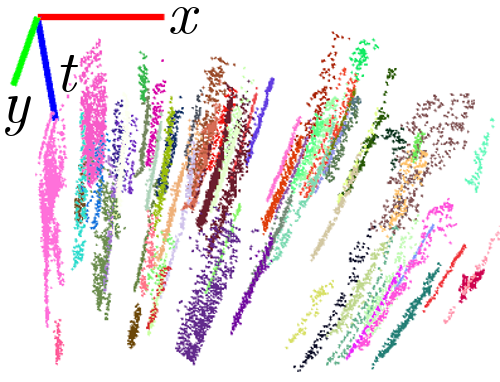}
        &
        \includegraphics[clip, width=\colsize,trim=0cm 0cm 0cm 0cm]{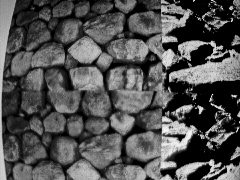}
        \\
        \hline
    \end{tabular}
    \includegraphics[clip, width=7.0cm,trim=0cm 0cm 0cm 0cm]{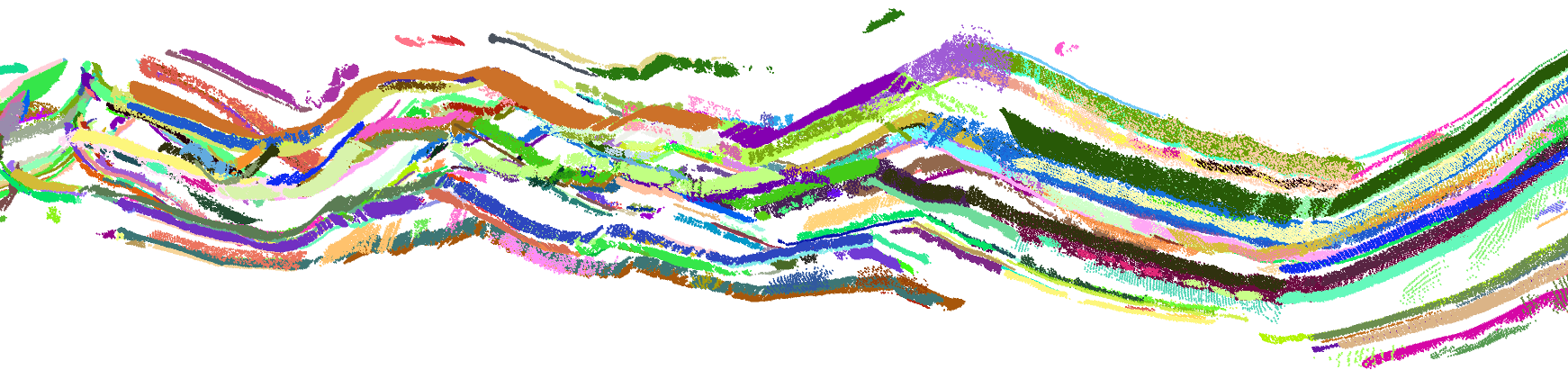}
    \caption{Examples of line clusters extracted in different dataset. The first two rows (raw event data, line clusters viewpoint A, and B, corresponding greyscale image) correspond to windows of 200k events. Bottom is cluster extraction on a very large window of 2M events in \texttt{shapes_6dof}.}
    \label{figure:frontend}
\end{figure}

The actual clustering is applied in a region growing fashion inspired by the connected component segmentation implemented in \cite{Zampogiannis2018}.
Colinearity of the \edit{$x$-$y$} normal vectors ($\normal{\i}{}^\top\normal{\j} > \normalthr$) and the point-to-line distance in the image space ($\lvert\normal{\i}{}^\top(\event{\j} - \event{\i})\lvert < \ptolthr$) are  used as the similarity criteria to assert that two neighbouring points belong to the same cluster.
Figure~\ref{figure:teaser} and \ref{figure:frontend} show examples of event clusters in the three types of datasets used in Section~VI.
\edit{By considering only the $x$-$y$ normals with a region growing algorithm, the proposed method allows for line clustering in a large variety of scenarios that are not restricted by the nature of the system's motion (e.g. constant velocity).}

The segment association between two consecutive windows is conducted by appending the last events of a window to the beginning of the following one and using the same similarity criteria described for the in-window clustering.
Each connected segment is attached to a line $\linestate{\world}{\l}$ and the event-line associations $\asso = \{ \event{\i}, \eventtime{\i}, \linestate{\world}{\l}\}$ are pushed to the set $\assoset$.
Note that, in the current implementation, the event-to-line association is only performed at the front-end level. There is no additional strategy to associate new segments to existing line estimates.
This would greatly improve the accuracy and robustness of the proposed method in scenarios where a same line results in multiple clusters that cannot be matched via the aforementioned procedures, or when lines reappear in the field of view after having left it.

\section{Experiments} \label{sec:experiments}

Our implementation uses \textit{Ceres}\footnote{http://ceres-solver.org/} for the non-linear least-square optimization of Equation~\eqref{eq:costfunction}.
At the current stage of development, \ac{idol} is computationally expensive as the full batch optimization is conducted every time a new event-window is processed.
\edit{The rapidly growing number of residuals} leads to prohibiting  \edit{estimation time of the system's Hessian}.
We arbitrarily chose to compute the full batch optimization until $\eventtime{} = 24\,\mathrm{s}$.
\edit{Intuitively, increasing the length of the full-batch time-window robustifies the system with respect to front-end failures.}
Then we switch to a sliding window optimization over the last $30$ event-windows with the last marginalized pose estimate fixed.

Due to the lack of publicly available event-based \ac{vio} algorithms, ROVIO \cite{Bloesch2015} was chosen as a comparison. ROVIO is a light-weight \ac{vio} algorithm that operates on traditional intensity images and is built upon an \ac{ekf} back-end. The front-end of ROVIO extracts patch-based features that are matched based on a photometric error resulting in a semi-dense approach. The system has been shown to be very robust, even under very aggressive motions \cite{Bloesch2015}.\\
Since \ac{idol}, in contrast to ROVIO, performs batch optimization, for a fair comparison we also included results obtained with maplab \cite{Schneider2017} by building a \ac{vi} pose-graph using ROVIO and jointly \edit{optimizing} the trajectory and sparse BRISK landmarks \cite{leutenegger:ICCV2011:brisk} using the maplab toolbox.

\subsection{Datasets and Evaluations}
In order to test \ac{idol} in challenging conditions and evaluate the performance compared to state-of-the-art in robust traditional \ac{vio}, tests on multiple real-world datasets including aggressive motions of the Event Dataset and Simulator \cite{Mueggler2017} were performed. The Event Dataset contains sensor data from a DAVIS240~\cite{Brandli2014} including the event data, traditional grey-scale images and \ac{imu} measurements. In particular, the indoors scenarios \texttt{shapes}, \texttt{poster} and \texttt{dynamic} were chosen as they include  6-\ac{dof} ground-truth from an indoor positioning system.
Both translation-only and full 6-\ac{dof} versions of these datasets have been used.
Rotation-only datasets were omitted since translation is necessary for good observability of the scene depth \cite{zhu:CVPR2017:evio}. 
\edit{In its current state, our front-end does not offer enough robustness to address the \texttt{boxes} scenarios of \cite{Mueggler2017} because of their very high level of texture in the scene.}
In addition to reporting qualitative results of the state progression and \acp{rmse} of aligned trajectories, the evaluation is also performed using a trajectory-segment based approach \cite{Zhang2018AOdometry} with segment lengths corresponding to $\unit[\{10, 20, 30, 40, 50\}]{\%}$ of the trajectory length.

\subsection{Results}
\begin{figure}
    \centering
    \includegraphics[width=0.9\columnwidth]{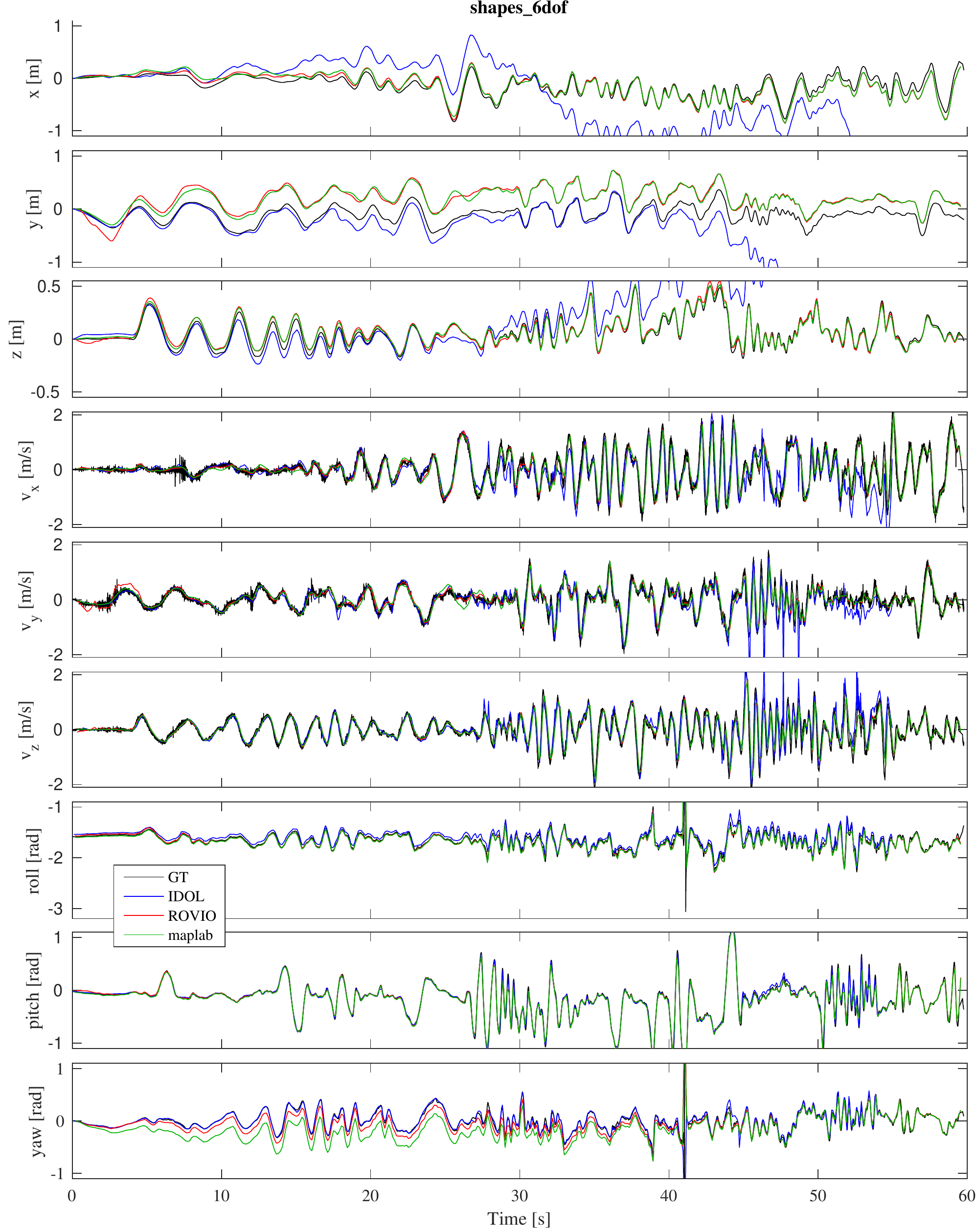}
    \caption{Pose and velocity estimates' progression of the different algorithms on the \texttt{shapes_6dof} dataset.}
    \label{fig:seperate_shapes_6dof}
\end{figure}

\begin{figure}
    \centering
    \includegraphics[width=0.9\columnwidth]{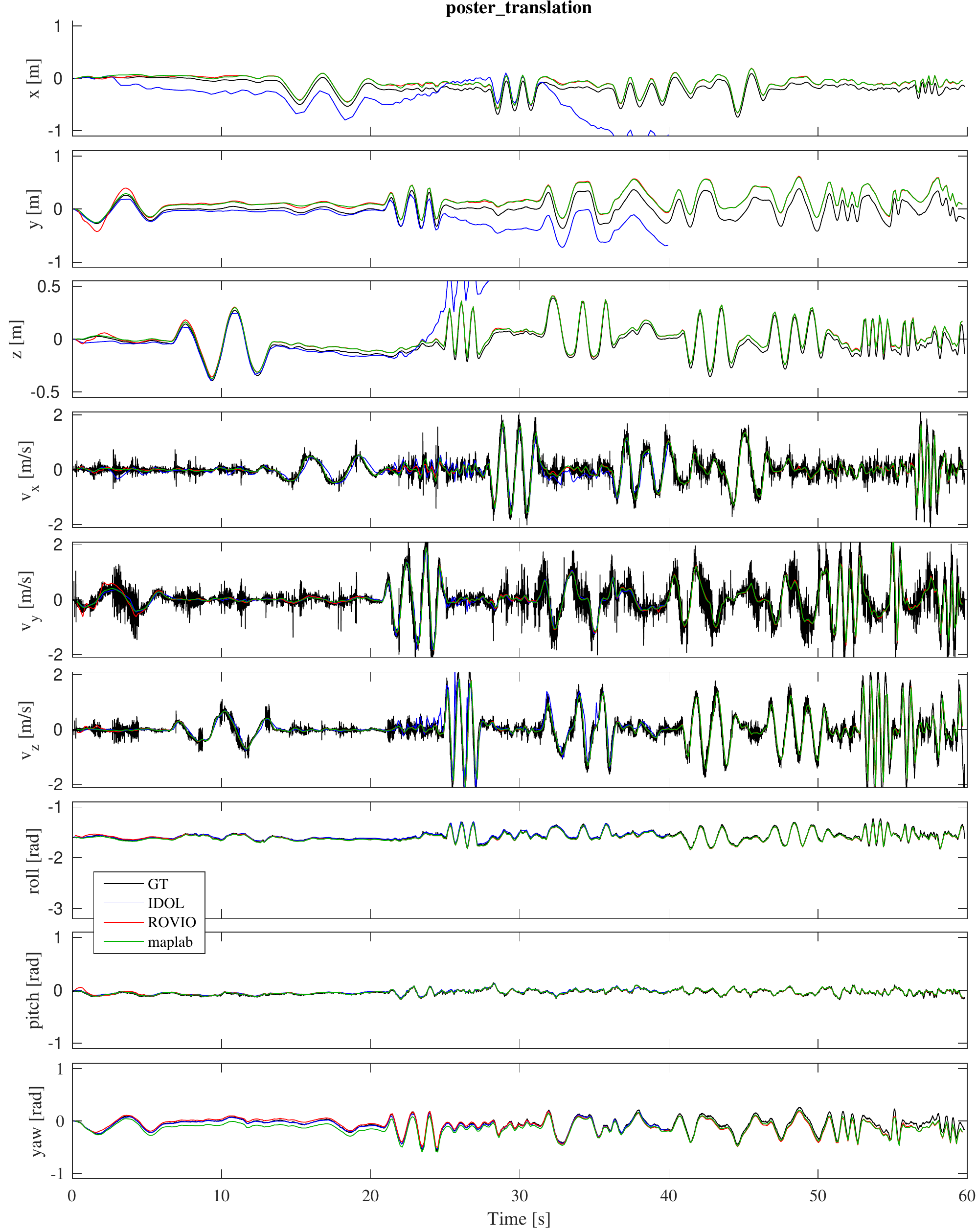}
    \caption{Pose and velocity estimates' progression of the different algorithms on the \texttt{poster_translation} dataset.}
    \label{fig:seperate_poster_translation}
\end{figure}

\begin{figure}
    \centering
    \includegraphics[width=0.9\columnwidth]{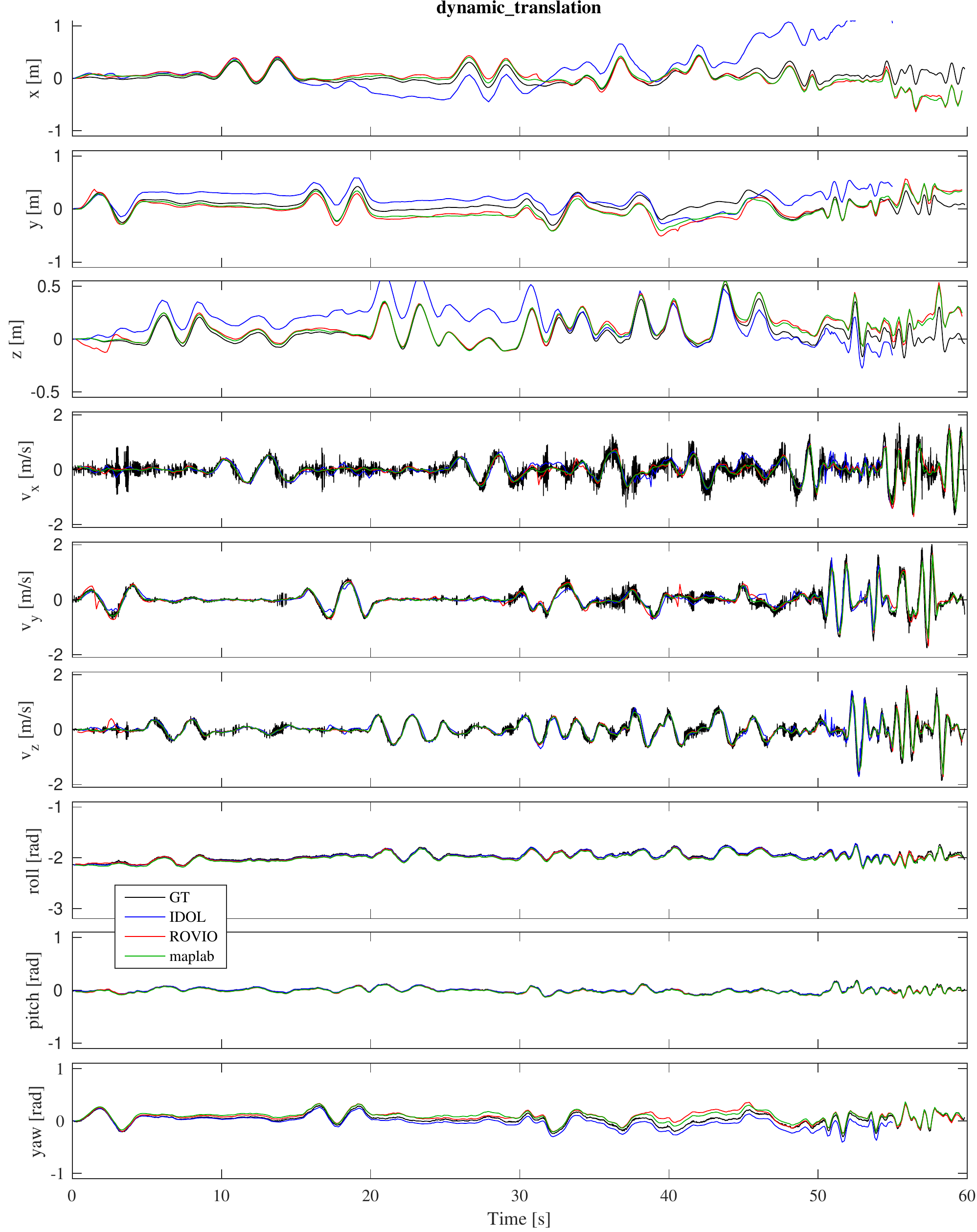}
    \caption{Pose and velocity estimates' progression of the different algorithms on the \texttt{dynamic_translation} dataset.}
    \label{fig:seperate_dynamic_translation}
\end{figure}

\begin{figure}
    \centering
    \includegraphics[width=0.85\columnwidth]{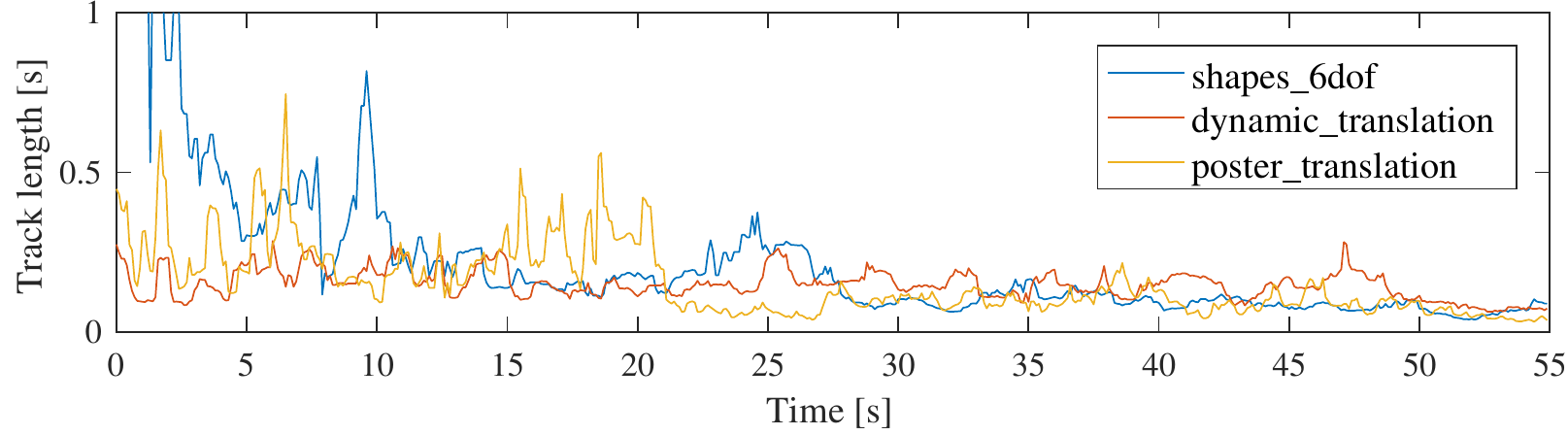}
    \caption{Average line-track lengths (moving average) across different datasets.}
    \label{fig:track-lengths}
\end{figure}

\begin{figure}
    \centering
     \subfloat[\texttt{shapes_6dof}] {%
        \includegraphics[width=0.7\columnwidth]{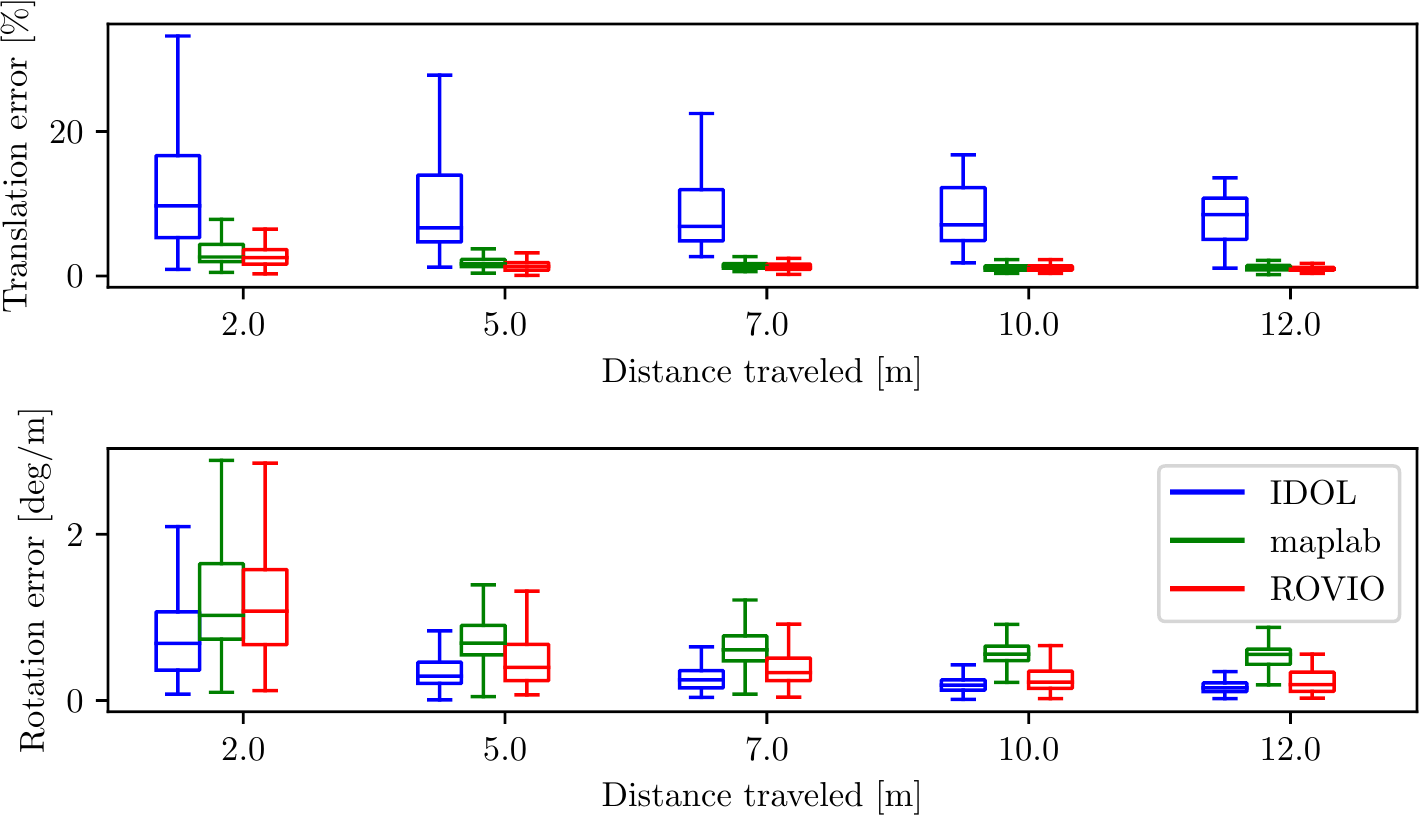}
        \label{fig:shapes_6dof_kitti}
        } \\
    \subfloat[\texttt{poster_translation}] {%
        \includegraphics[width=0.7\columnwidth]{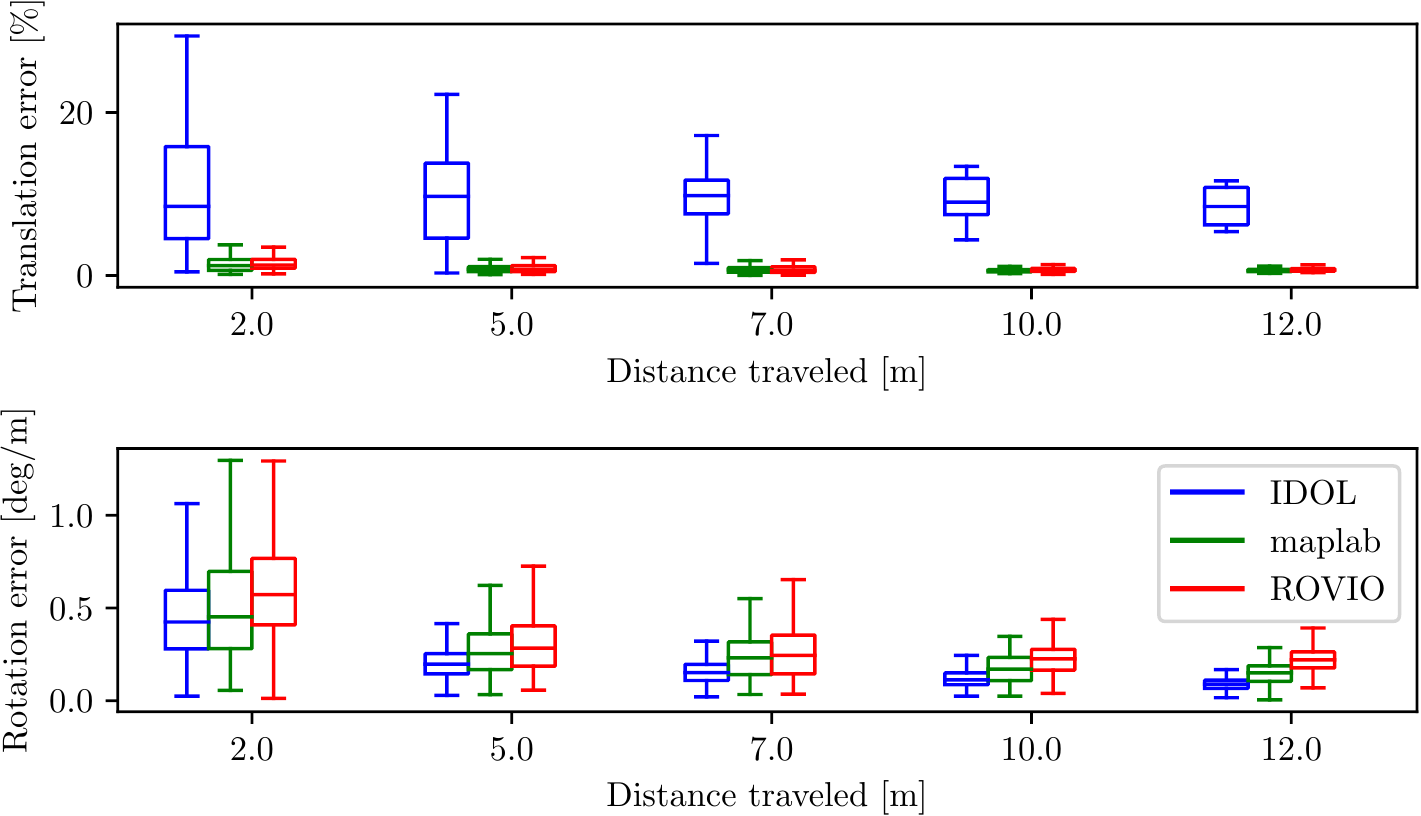}
        \label{fig:poster_translation_kitti}
        } \\
    \subfloat[\texttt{dynamic_translation}] {%
        \includegraphics[width=0.7\columnwidth]{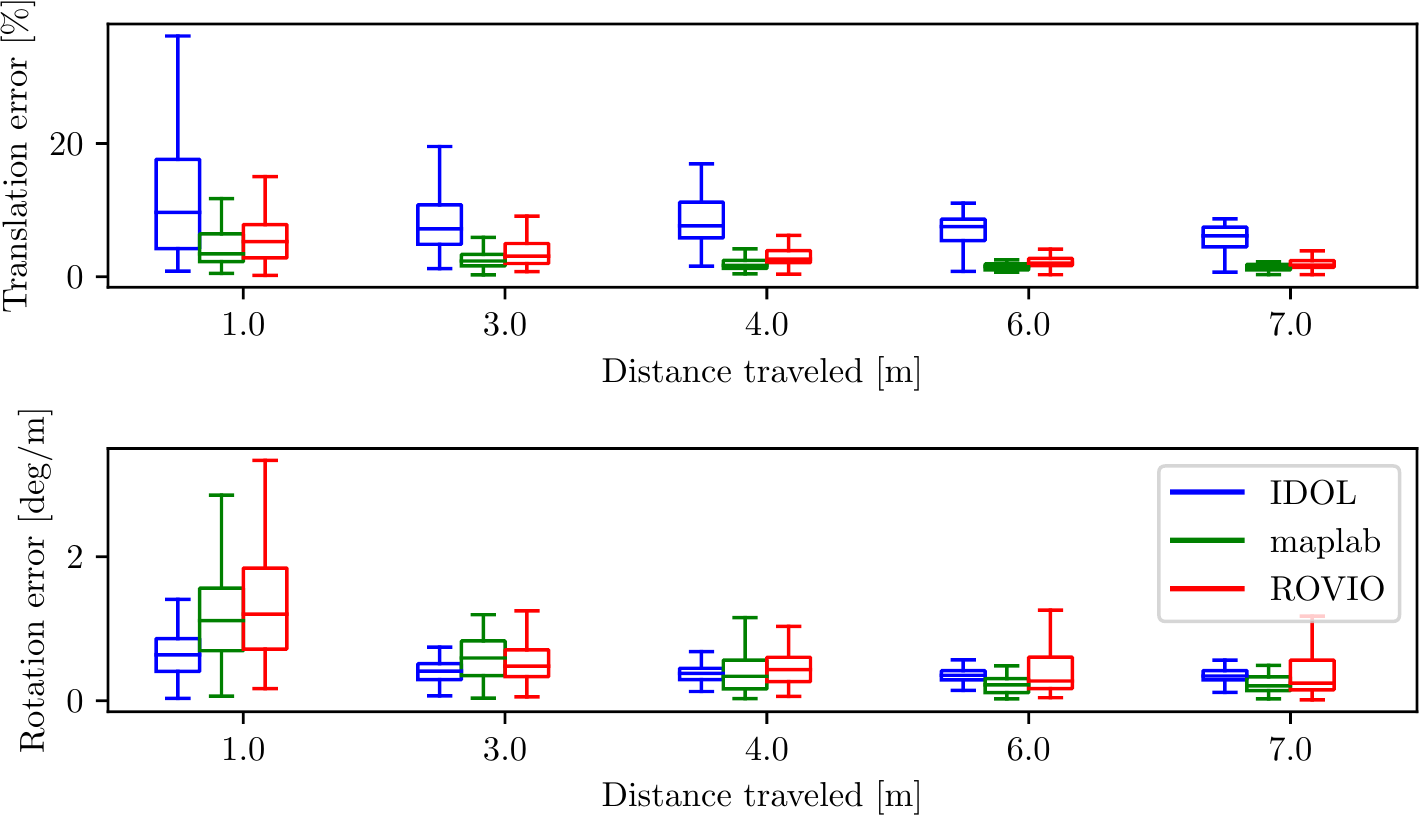}
        \label{fig:dynamic_translation_kitti}
        }
    \caption{Translation and orientation error of the first $\unit[40]{s}$ using the different algorithms upon different segment lengths.}
    \label{fig:kitti}
\end{figure}

Figures~\ref{fig:seperate_shapes_6dof}-\ref{fig:seperate_dynamic_translation} depict the state progression of the ground-truth and estimations of \ac{idol}, ROVIO and ROVIO+maplab on \texttt{shapes_6dof}, \texttt{poster_translation} and \texttt{dynamic_translation}. It becomes visible that, even though there is some drift in the translation estimation, \ac{idol} is able to estimate the camera pose and velocity, and especially the camera rotation, with high accuracy.  \\
After the initial stages of each of the experiments, we observe that the proposed \ac{vio} pipeline's translation estimate tends to diverge, which can be explained as a combination of different factors.
One can see that the state variables are generally well estimated during the full-batch optimization that takes places at the initial instants of each experiment and only starts drifting seconds after switching to the sliding-window mode. 
Our implementation could benefit from leveraging the uncertainty of the estimated pose across consecutive windows of optimization, properly marginalizing previous states.
We must also consider that, in its current form, the front-end described in Section~\ref{sec:frontend} produces rather short line-tracks. Figure~\ref{fig:track-lengths} depicts the average track length across the datasets.
One can observe the correlation between the drop of the track length around $\eventtime{} = 27\,\mathrm{s}$ and $\eventtime{} = 24\,\mathrm{s}$ in \texttt{shapes_6dof} and \texttt{poster_translation}, respectively, and the sudden increase of the translation errors.
Short track lengths and the absence of a strategy to re-detect previously observed lines do not account for a good depth estimation due to the low parallax.
Consequently, the translation estimates are notably affected whereas the rotation estimates, not as dependent on the depth estimate of the scene, still perform in a competitive range for small increments.
Note that despite a growing drift of the translation estimates likely due to our front-end's weaknesses, \ac{idol} performs accurate velocity estimation all along the trajectory validating its effectiveness for \ac{vio}.
Figure~\ref{fig:track-lengths} also shows that our front-end does not perform equally across the different datasets as per the different levels of noise in event data (due to high texture scenes) and the absence of noise filtering strategy.
Table~\ref{tab:rmse} reports \ac{rmse} of translation and rotation estimation of all algorithms on each of the datasets. The metric is highly depending on the translation estimate as the alignment method used for \ac{vio} minimizes the translation error to find a position and yaw offset for the whole trajectory.
Accordingly, if even only one component of the translation estimate diverges, the orientation \ac{rmse} is highly influenced by the poor trajectory alignment.
Consequently, the results depicted in Table~\ref{tab:rmse} are obtained by aligning only the $50$ first poses of the estimation and only taking the first $\unit[40]{s}$ into account. Typically, \ac{idol} performs worse than ROVIO and ROVIO+maplab in terms of translation estimation but still on the same order of magnitude, especially on \texttt{shapes_6dof} and \texttt{dynamic_translation}. Moreover, competitive results in orientation estimation can be achieved for most datasets. Only \texttt{shapes_translation} and \texttt{poster_translation} show a worse rotation estimation compared to ROVIO mainly due to diverged translation estimation (see Figures~\ref{fig:seperate_shapes_6dof}-\ref{fig:seperate_dynamic_translation}). \\
\begin{table*}[]
    \centering
    \begin{tabular}{l|cccc|cccc|cccc}
& \multicolumn{4}{c}{\texttt{shapes}}& \multicolumn{4}{c}{\texttt{poster}}& \multicolumn{4}{c}{\texttt{dynamic}}\\
& \multicolumn{2}{c}{\texttt{6dof}}& \multicolumn{2}{c}{\texttt{translation}}& \multicolumn{2}{c}{\texttt{6dof}}& \multicolumn{2}{c}{\texttt{translation}}& \multicolumn{2}{c}{\texttt{6dof}}& \multicolumn{2}{c}{\texttt{translation}}\\
&$e_t$&$e_r$&$e_t$&$e_r$&$e_t$&$e_r$&$e_t$&$e_r$&$e_t$&$e_r$&$e_t$&$e_r$\\ \hline \hline
\ac{idol}&0.52&8.35&0.51&18.62&0.62&6.15&0.70&10.82&0.54&6.08&0.25&4.92\\
ROVIO&0.34&10.24&0.05&1.56&0.15&8.21&0.13&2.09&0.16&10.27&0.12&6.19\\
ROVIO + maplab&0.25&12.91&0.05&2.25&0.12&3.53&0.09&1.92&0.08&6.09&0.09&3.15\\
    \end{tabular}
    \caption{Overall \ac{rmse} of translation estimation $e_t \, [ \unit{m} ]$ and orientation estimation $e_r \, [ \unit{deg} ] $ for \ac{idol}, ROVIO and ROVIO+maplab on different datasets ($\unit[0-40]{s}$, aligning first 50 poses). }
    \label{tab:rmse}
\end{table*}
More insights into the alignment issue can be found in the segment based evaluation shown in  Figure~\ref{fig:kitti}. It becomes clear that, in its current state, \ac{idol} can outperform both ROVIO and ROVIO+maplab in incremental rotation estimation but mainly lacks in translation estimation.

The results displayed above demonstrate the ability of \ac{idol} to perform \ac{vio} in various real-world scenarios.
Overall, the proposed method performs in the same order of magnitude than the compared frame-based methods according to the presented results. However, note that, at submission time, we are using an unrefined implementation of the event-based front-end, in which no noise filtering or re-detection scheme is applied. For instance, the knowledge of the line extremities' position could help the data association in scenarios in which the camera re-observes previously mapped areas. Additionally, no protection against outliers has been placed in the front-end nor in the back-end. In this regard, substantial improvements in performance are to be expected from the adoption of more robust strategies from the event-based and frame-based \ac{vio} literature \edit{with, for example, the adoption of a robust loss function.}

\section{Conclusions} \label{sec:conclusions}

This paper introduced \ac{idol}, a novel pipeline for event-based \ac{vio} using lines as features.
Unlike most of the event-based methods in the literature, the proposed method does not aggregate events into frames that are later used in a traditional frame-based \ac{vio} pipeline.
Here, the events are considered individually as part of a batch optimization that estimates the position of 3D lines alongside the system's pose and velocity.
This framework leverages the \acp{gpm} to characterize the system's trajectory based on a continuous representation of the inertial data.\\
We demonstrated the feasibility of event-based \ac{vio} with lines in different real-world scenarios.
While the presented system does not have real-time capabilities yet, \ac{idol} results show the potential to become an efficient, robust, and accurate event-based \ac{vio} method.
Across quantitative benchmarking against state-of-the-art frame-based \ac{vio} algorithms, \ac{idol} demonstrated accuracy in the same order of magnitude for translation, and competitive results for orientation.

Future work includes the exploration of different optimization strategies and the implementation of a probabilistic marginalization of the past state variables to substitute for the current growing full-batch optimization.
The computational cost of \ac{idol} can be addressed at different levels.
For example, one could optimize the implementation by using \ac{gpu} computation as most of the operations are highly parallelizable (normal estimations for each of the events in their spatio-temporal representation, computation of the residuals and Jacobians, inference of the per-event \acp{gpm}, etc.).
At a higher level, reducing the size of the optimization problem would greatly benefit the method's efficiency.
We will investigate different strategies to reduce the number of residuals while not sacrificing the amount of information used in the state estimation.
\edit{Finally, combining our line-based features with corner-based features is likely to improve both robustness and accuracy.}

\bibliographystyle{IEEEtran}
\bibliography{bibliography_florian,bibliography_cedric,bibliography_ignacio}

\begin{acronym}
\acro{imu}[IMU]{Inertial Measurement Unit}
\acro{gps}[GPS]{Global Position System}
\acro{gnss}[GNSS]{global navigation satellite system}
\acro{rtk}[RTK]{real time kinematics}
\acro{rmse}[RMSE]{root mean squared error}
\acro{hdr}[HDR]{High Dynamic Range}
\acro{slam}[SLAM]{Simultaneous Localization And Mapping}
\acro{dvs}[DVS]{Dynamic Vision Sensor}
\acro{ekf}[EKF]{extended Kalman filter}
\acro{kf}[KF]{Kalman filter}
\acro{vi}[VI]{visual-inertial}
\acro{vio}[VIO]{visual-inertial odometry}
\acro{vo}[VO]{Visual Odometry}
\acro{dso}[DSO]{direct sparse odometry}
\acro{dof}[DoF]{degree of freedom}
\acro{uav}[UAV]{unmanned aerial vehicle}
\acro{tof}[ToF]{time of flight}
\acro{lidar}[LiDAR]{light detection and ranging sensor}
\acro{mcu}[MCU]{microcontroller unit}
\acro{usb}[USB]{universal serial bus}
\acro{lidar}[LiDAR]{Light Detection and Ranging sensor}
\acro{spi}[SPI]{Serial Peripheral Interface}
\acro{ae}[AE]{auto-exposure}
\acro{gpu}[GPU]{graphic processing unit}

\acro{mav}[MAV]{micro aerial vehicle}
\acro{idol}[IDOL]{IMU-DVS  Odometry  with Lines}
\acro{gp}[GP]{Gaussian Process}
\acro{gpm}[GPM]{Gaussian preintegrated measurement}
\end{acronym}

\end{document}